\pdfoutput=1

\documentclass[11pt]{article}

\usepackage{acl}

\usepackage{times}
\usepackage{latexsym}

\usepackage[T1]{fontenc}

\usepackage[utf8]{inputenc}
\usepackage{times}
\usepackage{latexsym}
\usepackage{url}
\usepackage{enumitem}
\usepackage{amsmath,amssymb,amsfonts}
\usepackage{algorithmic}
\usepackage{caption}
\usepackage{subcaption}
\usepackage{graphicx}
\usepackage{textcomp}
\usepackage{url}
\usepackage{makecell}
\usepackage{multirow}
\usepackage{booktabs}
\usepackage{array}
\usepackage{microtype}
\usepackage{kotex}

\usepackage{multirow}
\usepackage{booktabs}
\usepackage{hyperref}

\usepackage{hyperref}       
\usepackage{amsfonts}       
\usepackage{nicefrac}       
\usepackage{xcolor}         

\newcommand{\thickhline}{%
    \noalign {\ifnum 0=`}\fi \hrule height 1pt
    \futurelet \reserved@a \@xhline
}

\usepackage{comment}
\usepackage{todonotes}
\usepackage{xcolor}

\usepackage{microtype}
\title{Self-Improving-Leaderboard (SIL): A Call for Real-World Centric Natural Language Processing Leaderboards}

\author{Chanjun Park$^{1,2}$, Hyeonseok Moon$^{1}$, Seolhwa Lee$^{3}$, \\ {\bf \large Jaehyung Seo$^{1}$, Sugyeong Eo$^{1}$,  Heuiseok Lim$^{1,\dagger}$\thanks{$\dagger$ Corresponding author}}\\
\\
  $^1$Korea University, $^2$Upstage ,$^3$Technical University of Darmstadt\\
  \texttt{\{bcj1210, glee889, seojae777, whiteldark, djtnrud, limhseok\}@korea.ac.kr} \\
  \texttt{chanjun.park@upstage.ai}
  }

\begin{document}
\maketitle
\begin{abstract}
Leaderboard systems allow researchers to objectively evaluate Natural Language Processing (NLP) models and are typically used to identify models that exhibit superior performance on a given task in a predetermined setting. However, we argue that evaluation on a given test dataset is just one of many performance indications of the model. In this paper, we claim leaderboard competitions should also aim to identify models that exhibit the best performance in a real-world setting. We highlight three issues with current leaderboard systems: (\romannumeral 1) the use of a single, static test set, (\romannumeral 2) discrepancy between testing and real-world application (\romannumeral 3) the tendency for leaderboard-centric competition to be biased towards the test set. As a solution, we propose a new paradigm of leaderboard systems that addresses these issues of current leaderboard system. Through this study, we hope to induce a paradigm shift towards more real -world-centric leaderboard competitions. 
\end{abstract}

\section{Introduction}
Leaderboard systems allow researchers to compete with artificial intelligence (AI) models. The general purpose of leaderboard competition is to identify models that exhibit superior performance on a given task~\cite{liu2021explainaboard, mazumder2022dataperf}. Most of these competitions focus on solving real-world task, such as a machine reading comprehension \cite{rajpurkar2016squad} or a question-answering task \cite{lai2017race}. As a representative of the real-world task, most of each current leaderboard system propose a test dataset that serves as a measure of model performance for the corresponding task \cite{park2021klue, wang2018glue}. 

However, we should investigate that whether the current leaderboard competition system sufficiently considers the intended goal. While most leaderboard systems use a single test set for the model evaluation in each task~\cite{rajpurkar2016squad, wang2018glue}, this may derive biased evaluation~\cite{beyer2020we, van2021we}. In particular, in utilizing a single-static test dataset, we are exposed to various sources of bias, including annotation bias~\cite{gururangan2018annotation, hovy2021five}
 
We argue that this setting may derive the leaderboard competition to be recognized as a race for the higher rankings rather than the consortium for developing better models for the objective real-world task~\cite{van2021we}. 

In this paper, we extend this discussion to analyze the differences in the current NLP leaderboard ranking system from the perspective of real-world applications in the following three aspects.

First, several leaderboards adopt static test dataset \cite{wang2018glue, farhad2021findings}. Hundreds of millions of data points are generated every day in the real world \cite{agababov2015flywheel, ramkumar2019artificial, liu2022real}. In several industrial services, data are generated and evolved daily, and the newly gained user history acts as the key to the model performance \cite{resnick1997recommender, shin2021scaling}. However, we find that existing leaderboards are stuck with outdated data and do not reflect the current situation. In other words, while real-world data are dynamic and continuously accumulating, leaderboard data remain static.

Second, current leaderboards hardly reflects real-world environments \cite{raji2021ai, thompson2020meta}. As data are continuously accumulated via several sources, such as users or industries, we can find that edge cases and outliers continue to occur~\cite{lyu2020revisiting}. In particular, errors, typos, and noisy data always exist in the industrial services, which require models that are sufficiently flexible to deal with them. The current leaderboard system, which operates only on clean data, does not have a way to measure this flexibility in models \cite{loic2020findings}. Optimal models should be able to adapt to real-world data that is continuously generated, changing, and evolving~\cite{grispos2019good, budach2022effects}.

Finally, existing systems are leaderboard-centric. For instance, several models were adjusted to the test set to correspond to the leaderboard~\cite{zhang2017position, khashabi2021genie}. However, this approach can accelerate overfitting, resulting in test set fitted-only model while gives deteriorated performance on the real-world data. In other words, the current leaderboard system is not operated in a real-world centric manner but rather in a leaderboard-centric manner. Thus, we have to determine whether the current system of competition is meaningful

To address these issues, we propose a new paradigm for leaderboards: the Self-Improving Leaderboard (SIL). SIL is based on a continuously self-improving test dataset, with rankings updated every 24 hours according to the data of that day. The key feature of this leaderboard is that the rankings are constantly updated while the model remains fixed. In other words, this concept addresses the inherent flaws of conventional leaderboards by retaining the model characteristics and permitting the data to evolve. 

\section{Hurdles to developing Real-World-Centric Leaderboards}
In this section, we analyze current leaderboard competitions and identify their limitations from the perspective of real-world environment. In particular, three challenges are identified.

\paragraph{Static Dataset}
Most leaderboards and benchmark data have been tested and verified by numerous models over a long period of time since their release. However, it is important to note that these data may not reflect the changes that occur in real-world scenarios. In practice, systems that constantly update models based on new data, such as data flywheel systems and continual learning methods, are often used to keep models current and improve their performance~\cite{kreuzberger2022machine}.

As various services are provided in the real-world, new data is constantly generated and the model is continuously improving through continual learning, creating a structure called a data flywheel~\cite{agababov2015flywheel}. While many companies are adopting this process. we argue that current leaderboard systems are not able to consider this dynamic characteristics of data.

In other words, current leaderboard systems do not reflect these data-flywheel systems as whole evaluation is performance based on a single test set \cite{van2021we}. This limits the objective evaluation on the practical usability of a model in an actual setting. 

\paragraph{Discrepancy between testing and real-world application}
When evaluating the performance of a model, most of them are evaluated on clean data as a common practice. The purpose here is to measure the performance in a state where errors or biases present in the test data are removed as much as possible. However, this may be a suitable method for evaluating excellence in targeting tasks, but it can be difficult to evaluate whether the model can produce robust performance in real-world settings where various errors exist.

In other words, data without any errors or biases is often referred to as ``gold'' data, and using it to evaluate a model's performance is considered to be the most objective way to measure its capabilities~\cite{gehrmann2021gem, koo2022k}. Hence, the quality of data is regarded in all areas of NLP, and  methods for improving data to meet specific needs are actively studied~\cite{koehn2020findings}.

However, when considering real-world applications, this strict focus on clean text may be considered as an additional bias, since even real-world data is not cleaned~\cite{liu2022towards}. Currently, most NLP models are trained using clean texts in general \cite{petroni2020kilt, wang2019superglue}, while real-world setting include noisy or inconsistency data \cite{stickland2022robustification}. This may exacerbates the gap between the data on which the model performs well and the data on which the model should perform. We argue that models need to be robust to noise to actually solve--- ``the problems that the dataset aims to solve''~\cite{wang2020comprehensive}. 

From this perspective, test data in current use are not as suitable for validating the performance of real-world centric models. In other words, there can be a discrepancy between the performance of a model when it is evaluated on clean data during testing and its performance when it is applied in the real-world.

\paragraph{Leaderboard-Centric Competition}
As leaderboards operate over longer periods, the number of participating models increases. However, the emphasis is placed solely on evaluating performance of each model on the given test dataset, neglecting performance in other environments or test sets. As a result, the original goal of the competition, which is to develop models that can effectively address real-world challenges, is obscured by the emphasis on achieving high performance on a single test dataset.

In particular, earlier studies have highlighted that employing a fixed single test dataset are extremely vulnerable to biases inherent in NLP application~\cite{hovy2021five}. As the number of competition trials using such limited data increases, studies on the leaderboard are likely to overfit the proposed test dataset, not the corresponding task itself~\cite{rodriguez2021evaluation, rosenman2011measuring}. We should note that the testset is selected as a representative of the given task. Adoption of test dataset in evaluating each model is not problematic \cite{lee2019use}, but it becomes complicated in the leaderboard competition where the verification work should indispensably be repeated several times \cite{van2021we}. 

Furthermore, existing works suggest more severe situations including annotation artifacts that the data selected to constitute the test set cannot sufficiently reflect objective of the intended real-world task~\cite{chen2020model, gururangan2018annotation}. Competitions based on such data fundamentally result in an imbalanced focus on the given static dataset \cite{rosenman2011measuring}. In other words, as the number of teams participating in ranking through static test sets increases, these biases are exacerbated, and all models are likely to overfit the publicly released static dataset as a benchmark~\cite{scargle1999publication}.

\section{Self-Improving Leaderboard (SIL)}
Considering the purpose of using the existing leaderboard ranking system, we propose that model performance evaluation should move beyond--- \textit{``model learning for the model''} and towards \textit{``model learning for real world applications.''} Accordingly, we propose three characteristics that would address the problems of the existing leaderboard system: an evolving test dataset, periodic ranking system, and comprehensive evaluation. Based on these characteristics, we introduce the SIL as an ideal leaderboard system, which is based on \textit{self-improving test datasets}.

\begin{figure*}[h!] 
\begin{center}
\includegraphics[width=\linewidth]{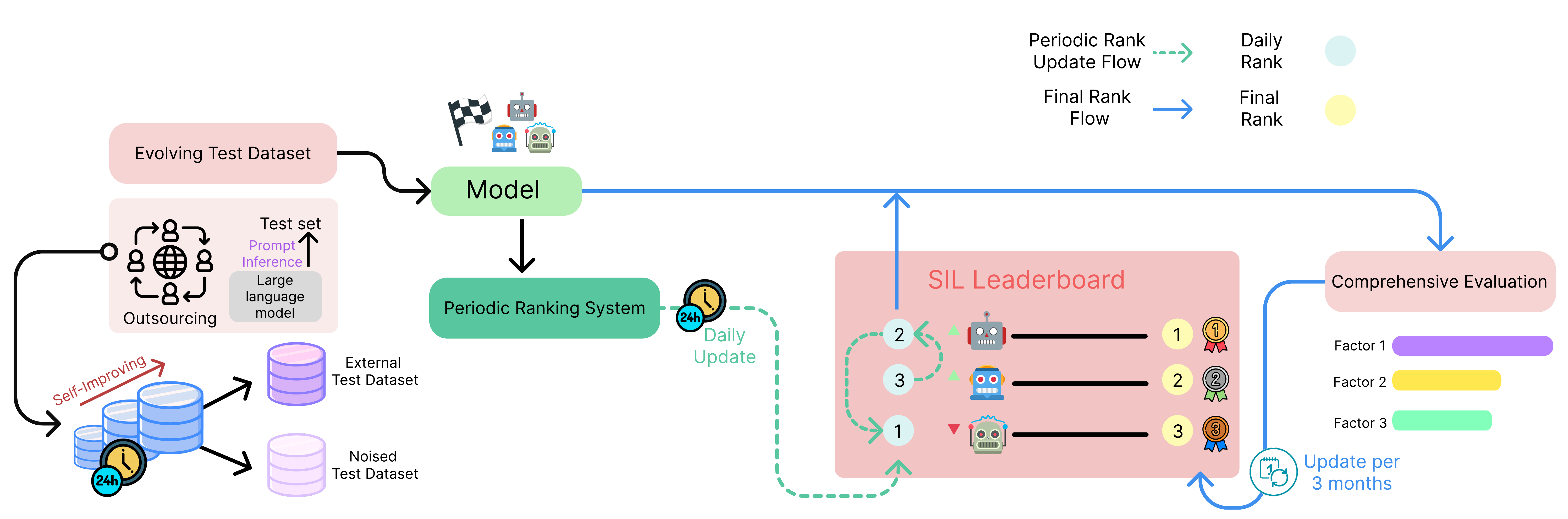}
\end{center}
\caption{Architecture of the Self-Improving Leaderboard (SIL). } 
\label{fig:overall}
\end{figure*} 

\subsection{Evolving Test Dataset}  
As previously mentioned, most models that have proven their capability through leaderboards suffer from excessive overfitting of task data \cite{roelofs2019meta, raji2021ai}. Despite being in the same domain and task, they do not perform well when tested in real-world scenarios~\cite{engstrom2020identifying}. Therefore, we propose to overcome this limitation by introducing a evolving test dataset that enables the evaluation of various aspects. We proposed two methods for data transformation and evaluation: an ``External Test Dataset'' and ``Noise Injection''.

\paragraph{External Test Dataset}
First, we propose to use an outsourced test dataset, i.e., an external source of test data for the target task. External sources include using task-related test data provided by individuals or companies in the verification phase or generated test data through prompts from large language models (LLMs) such as GPT3 \cite{yoo2021gpt3mix}. In particular, large language models, such as GPT3 and ChatGPT~\footnote{\url{https://chat.openai.com/}}, are currently being used in various fields with decent performance in NLP tasks, and are often show human-replaceable level of performance \cite{elkins2020can, wang2021want}. In this sense, evaluating model performance via distillation of the knowledge in a large language model can aid us in comprehensively evaluating various aspects \cite{meyer2022we}.

\paragraph{Noise Injected Test Dataset}
Real-world data always contain noise~\cite{yue2020dual}. Considering that, data integrity may not guarantee fair evaluation, but can deepen the gap between the evaluation and the real-world suitability \cite{zhou2018data}. We argue that evaluation with noised setting should be considered to match the evaluation environment with the real-world setting. We propose to separately evaluate each model participated in the leaderboard competition, with noised data. Through this process, we can estimate each model's robustness to simple changes as well as its ability to maintain its performance in unstable situations \cite{xie2017data}.

The method of injecting noise can be easily implemented by applying the approaches proposed in the fields of adversarial attack and spelling correction~\cite{morris2020textattack, park2021neural}. Additionally, only about 15\% of the original test data accumulated per day is augmented with noise injection. It is hard to measure the exact number of noise errors that occur in real-world NLP systems, since they may not always be easily spotted or reported. Therefore, the 15\% figure should be taken as an example rather than a precise measurement.

\subsection{Periodic Ranking System}
Over time, the evaluation data on leaderboards gradually become outdated \cite{beyer2020we}. Furthermore, as more competitors participate, the performance on the leaderboard hardly reveals the corresponding suitability to the real-world setting \cite{van2021we}. In alleviating these, we propose to set a regular performance update cycle for the evaluation data that considers the characteristics of constantly changing data. 

We judged that it would be reasonable to update the ranking every 24 hours, since new logs or data are accumulated every day when performing actual NLP services. Participants in the leaderboard, whose rankings are updated daily, will have to devise a new methodology beyond changing the model architecture (i.e. model-centric) or data-cleansing approach (i.e. data-centric~\cite{mazumder2022dataperf}) to improve their ranking. We believe that these considerations eventually promote real-world centric research.

Consequently, the dynamic capabilities of models can be evaluated through periodic evaluations, rather than determining their capabilities at the initial evaluation point \cite{beyer2020we}. The new ranking system also reflects instability and potential for changes in real-world data~\cite{thompson2020meta}.

\subsection{Comprehensive Evaluation}\label{sec:comprehensive}
In evaluating the performance of the model through multiple test datasets (i.e., evolving test dataset and periodic ranking system), we propose to eliminate the approach of determining the ranking of participating models based on a single performance or condition \cite{vajjala2022we}. Instead, we aim to obtain a comprehensive score for all validation environments to determine whether the model can maintain superior performance even when faced with data that may change constantly and include unexpected errors.

In summary, the testset and ranking change every 24 hours. However, the comprehensive evaluation is conducted quarterly (every 3 months). In other words, the final ranking is released every 3 months and the evaluation factors can include "how long the ranking has been maintained in the top position", "how small the range of ranking changes has been", and so on. It may also be possible to apply a medal system similar to Kaggle~\footnote{\url{https://www.kaggle.com/progression}} based on the final ranking.

In the end, our suggested procedure is not ended as a one-time ranking process (never-ending ranking system), unlike the existing leaderboard, while the final ranking is determined every quarter, the test set will continuously evolve, so the competition will not end. Medals will be awarded every three months, but the competition will continue indefinitely~\cite{mitchell2018never}. 

\subsection{Overall Process of SIL}
Our suggested procedure for the leaderboard ranking process is as follows in Figure~\ref{fig:overall}. 

Firstly, the evolving test dataset is supplied by the company or through the prompts using Large Language Model (LLM) to preserve the quality of the test dataset. In specific, the above test set is growing like self-improving every 24 hours cycle and is separated into an external test dataset and a noised test dataset, respectively (\textit{black colored direction}). 

Secondly, the periodic ranking system scores the daily rank of the models using an up-to-date test set (i.e., evolving test dataset) with daily cycles to check the model's robustness (\textit{green colored direction}). 

Thirdly, the final rank is derived quarterly (every three months) through comprehensive evaluation using the above daily ranks and many different factors, mentioned in Section.~\ref{sec:comprehensive} (\textit{blue colored direction}). 
Consequently, models are updated in SIL leaderboard dynamically.       

Our proposed SIL is real-world centric compared with existing leaderboards in the following ways. First, ``data are the most up-to-date''. In SIL, the data are maintained as the most updated version. Second, ``data are real-world similar''. SIL offers data that are not of high quality, but rather data that are consistent with actual situations in contrast to the noise-cleaned gold test sets. Lastly, ``data are not static and are subjected to less biases''. 

\section{Conclusions}
This paper is a position paper, discussed the limitations of conventional NLP leaderboards and proposed a new paradigm for leaderboard systems called SIL. Conventional leaderboards bears several challenges in real-world scenarios as they evaluate the performance of participating models solely with a single test dataset. We argue that this lead to even challenging estimation of model robustness and reproducibility in real-world environments.
Another problem is that competition accelerates overfitting, i.e., the model performs well only within the test set provided by the leaderboard. To address these problems, we proposed a new paradigm of leaderboard system called SIL, which continuously improves by updating the test data and reevaluating the models and rankings daily. This new type of leaderboard operates in a more real-world-centric manner rather than solely focusing on ranking up within the leaderboard. Hopefully, this work can trigger NLP community to apply this paradigm widely in real-world.

\section*{Limitations}
As a position paper, we present a new perspective on the leaderboard systems. While there is certainly room for additional empirical analysis, we argue that the insights provided in this work would serve as a starting step toward the new perspective. We contend that the novel perspectives on leaderboard systems presented in this paper will foster the emergence of leaderboards that are better suited to real-world centric and offer insight into the concept of valuable competition. 

With the Breakthrough emerging of ChatGPT, many researchers and people without expertise in NLP knowledge have been astonished by its abilities. Now they consider using ChatGPT in their real world life indeed, not just as an okay assistant. Likewise, our community is growing fastly, and we need to go through more real-world centric models and suitable cases for its evaluation.

\section*{Ethics Statement}
There are no ethical issues to declare.

\bibliography{anthology,custom}
\bibliographystyle{acl_natbib}

\end{document}